\documentclass[runningheads]{llncs}

\usepackage{graphicx}
\usepackage{booktabs}

\usepackage[accsupp]{axessibility}  

\usepackage{xcolor}  

\usepackage{algorithm}
\usepackage{algorithmic}
\usepackage{amsmath}
\usepackage{amssymb}

\usepackage{multirow} 

\usepackage{cite}

\usepackage{hyperref}
\usepackage[capitalize]{cleveref}

\usepackage{orcidlink}

\begin{document}

\title{SUPER Module for Detail-Sensitive and Cost-Efficient U-Net Variant Decoders} 

\titlerunning{SUPER Module}

\author{
    Siheon Joo\inst{1} \and 
    Hongjo Kim\inst{2}\thanks{Corresponding author}
}
\authorrunning{S.~Joo and H.~Kim}

\institute{
    Integrated M.S./Ph.D. Student, Dept. of Civil and Environmental Engineering, Yonsei Univ., Seoul, Korea\\
    \email{sh.joo@yonsei.ac.kr}, \url{http://sh-joo.github.io} 
    \and
    Associate Professor, Dept. of Civil and Environmental Engineering, Yonsei Univ., Seoul, Korea\\
    \email{hongjo@yonsei.ac.kr}, \url{http://hongjo.github.io} 
}
\maketitle

\begin{abstract}
Skip-connected U-Net variants dominate inverse problems, yet spatial upscaling can limit fine-structure preservation. While wavelet transforms offer high-frequency fidelity via Perfect Reconstruction (PR), this constraint is unnecessarily rigid when inputs and outputs differ. 
We propose \textbf{Selectively Suppressed Perfect Reconstruction (SUPER)}, a theoretical framework that relaxes PR into a learnable suppression operator.
By modularizing this into a \textbf{plug-and-play} decoder unit, SUPER bypasses conventional upscaling, explicitly disentangling signal preservation from nuisance suppression. 
Empirically, SUPER delivers a highly favorable \textbf{Pareto improvement} between computational cost and detail sensitivity. 
For detail-critical tasks, it robustly enhances edge recovery in depth estimation and sub-4px crack detection. 
Crucially, even under a severely degraded image denoising scenario—a rigorous stress test inherently lacking fine high-frequency cues—the SUPER Module substantially slashes decoder computational cost without sacrificing structural quality. 
These results validate SUPER as a highly practical plug-and-play solution, demonstrating that simple integration into existing U-Net variant decoders consistently improves both fine-detail accuracy and computational efficiency.
\end{abstract}

\begin{figure}[tb]
  \centering
  \includegraphics[width=\linewidth]{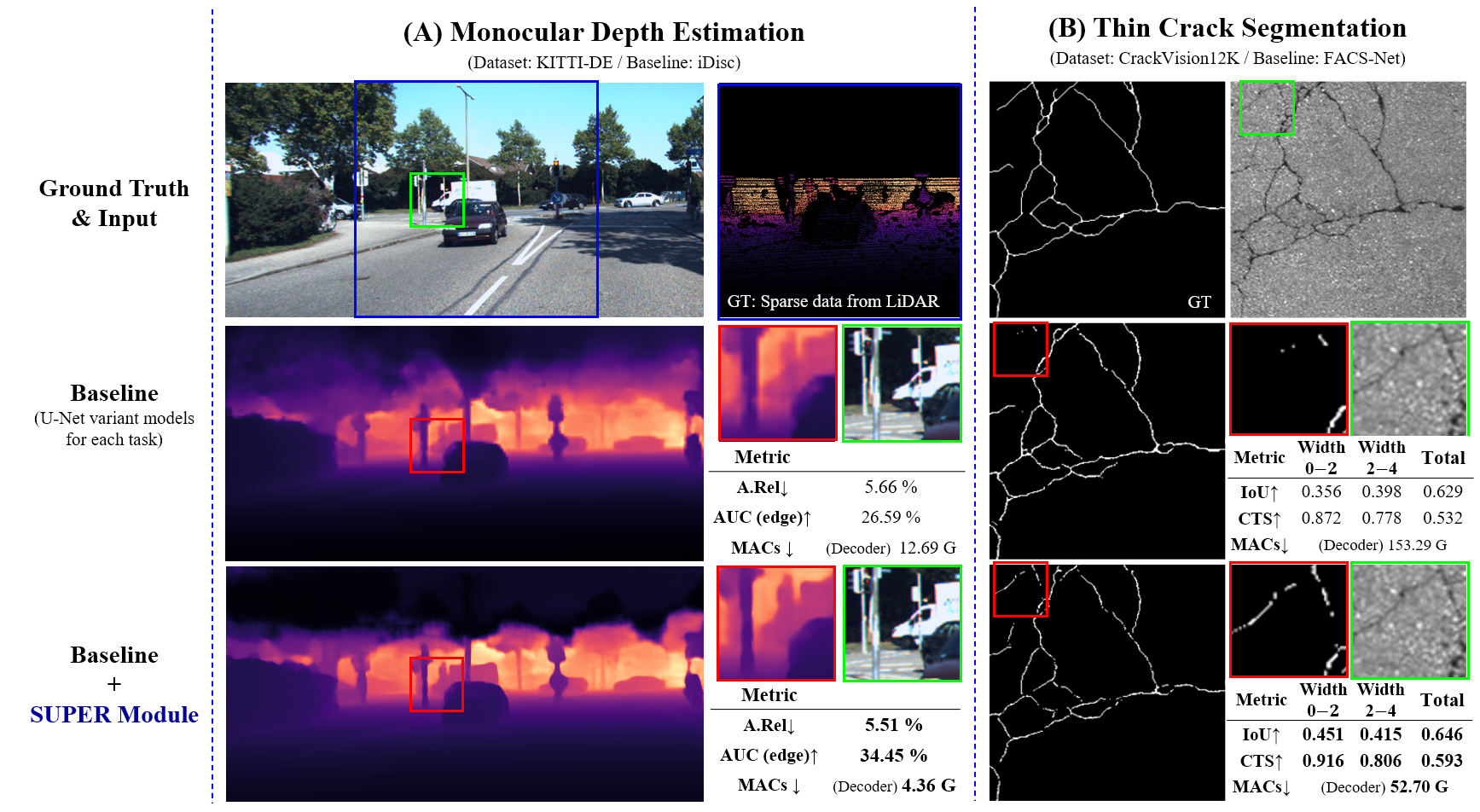}
  \caption{\textbf{Visual summary of the SUPER Module.} Our plug-and-play decoder module consistently improves high-frequency detail recovery while significantly reducing decoder MACs across diverse dense prediction tasks.}
  \label{fig:qualitative}
\end{figure}

\section{Introduction}
\label{sec1}
Recovering fine-scale structures---such as local anomalies or subtle textures---remains a fundamental bottleneck in skip-connected encoder--decoder architectures (often referred to as U-Net variants \cite{Ref04}). 
While these models dominate dense prediction tasks \cite{Ref01, Ref02, Ref05}, their decoders suffer from an intrinsic structural flaw. 
Specifically, after encoder downscaling discards fine details by reducing the Nyquist limit \cite{Ref03}, conventional decoders rely on spatial upscaling to restore resolution. 
However, this upscaling introduces artificial frequencies that blur true high-frequency structures, utilizing computational capacity inefficiently \cite{Ref06, Ref07}. 
Although skip connections attempt to reinject the lost cues, these signals are inherently sparse and low-energy \cite{Ref08, Ref09}. 
Compounded by the network's spectral bias \cite{Ref10, Ref11}, the decoder naturally favors fitting the low-variance artificial frequencies, overpowering true sparse details and yielding distorted reconstructions.
    
One promising direction to address these issues is leveraging wavelet transforms~\cite{Ref03}. 
The discrete wavelet transform (DWT) provides perfectly invertible downscaling, explicitly separating features into frequency bands—a property that inherently helps mitigate spectral bias~\cite{Ref12, Ref13, Ref14}. 
However, wavelet-based architectures are typically constrained by rigid conditions such as tight-frame constraints~\cite{Ref03}, enforcing strict analysis-synthesis relationships where the reconstruction filters must perfectly dual the decomposition filters~\cite{Ref17}.
This enforces a structural rigidity that prevents the network from learning task-specific feature transformations.
Consequently, conventional U-Net variants remain dominant~\cite{Ref19, Ref20, Ref21}, even though their high-frequency distortions degrade fine-detail accuracy---a limitation often masked by average-based evaluation metrics~\cite{Ref18, Ref22}. 
Suboptimal preservation of these high-frequency structures causes models to miss the thinnest cracks in segmentation~\cite{Ref23, Ref24}, overlook micro-lesions in medical imaging~\cite{Ref25}, or degrade boundary localization in depth estimation~\cite{Ref26}. 
Conversely, naively enforcing a high-frequency bias can amplify noise and destabilize low-frequency robustness~\cite{Ref14, Ref22}.

To address this trade-off, we propose \textbf{Selectively Suppressed Perfect Reconstruction (SUPER)}, a theoretical framework for frequency-selective feature restoration, which we modularize into a \textbf{plug-and-play SUPER Module} for U-Net decoders. 
Inspired by the conditional perfect reconstruction (PR) property of wavelets~\cite{Ref03}, SUPER selectively suppresses redundant features while retaining task-relevant high-frequency details. 
Free from tight frame constraints, it effectively bypasses standard spatial upscaling within the decoder. 
Evaluations on well-established, highly competitive baselines across monocular depth estimation (iDisc~\cite{Ref29}), thin-crack segmentation (FACS-Net~\cite{Ref24}), and single image denoising (CascadedGaze~\cite{Ref31}) demonstrate SUPER's general applicability. 
These experiments confirm its structural advantages: improving high-frequency fidelity, and, crucially, maintaining structural robustness in image denoising even while operating at a significantly reduced computational scale. This highlights SUPER as a Pareto-efficient solution that optimizes the balance between performance and computational.

\subsubsection{Key Contributions}
\begin{itemize}
\item We propose \textbf{SUPER}, a relaxed perfect reconstruction framework that brings wavelet-level precision to standard encoder--decoder architectures.
\item SUPER serves as a \textbf{plug-and-play} module that delivers \textbf{Pareto improvement}---enhancing fine-detail sensitivity while substantially reducing the computational cost of U-Net variant decoders.
\end{itemize}

\section{Related Works}
\label{sec2}

\subsubsection{The Importance of Details} in inverse problems is evident, as fine-scale structures such as edges, textures, and thin lines critically influence perceptual realism and task-specific accuracy across domains.
Prior surveys in video super-resolution~\cite{Ref01} and high-resolution image processing~\cite{Ref02} highlight that even minor structural degradations can impair interpretability, a problem echoed in crack detection~\cite{Ref23, Ref24}, medical imaging~\cite{Ref25}, depth estimation~\cite{Ref26}, and acoustic tomography denoising~\cite{Ref27}. 
Recent works address these issues through frequency-aware losses (e.g., FFL~\cite{Ref18}) and topology-aware constraints such as the Crack Topology Loss introduced in FACS-Net~\cite{Ref24}, which capture subtle distortions often overlooked by average-based metrics~\cite{Ref32}. 
Collectively, these studies emphasize the necessity of high-frequency preservation and the challenges faced by current architectures in ensuring it.

\subsubsection{U-Net Variants} remain the dominant design for detail-sensitive tasks due to their simplicity and strong feature reuse~\cite{Ref05}. 
However, their decoders face intrinsic challenges that can limit fine-detail recovery. First, standard spatial upscaling processes often distort structural information, producing blurred reconstructions even with skip connections~\cite{Ref04, Ref08, Ref33, Ref34, Ref35}. Second, while skip connections reinject high-frequency cues, the transmitted details are typically sparse and low-energy~\cite{Ref23, Ref25}. Third, neural networks exhibit spectral bias~\cite{Ref10}, preferentially fitting low-frequency components while underfitting high-frequency signals, exacerbating the difficulty of recovering subtle textures~\cite{Ref09, Ref11}. Fourth, many U-Net variants reduce computational cost by projecting fused features into fewer channels, but such compression risks discarding fine details~\cite{Ref36, Ref37}. Collectively, while widely adopted, U-Net variants remain structurally constrained in reconstructing high-frequency details.

\subsubsection{Wavelet-Based Architectures} 
have emerged as promising alternatives for fine-detail preservation. By exploiting the perfect reconstruction property of wavelet transforms~\cite{Ref03}, they prevent information loss during sampling~\cite{Ref06, Ref12} and explicitly separate features into frequency bands~\cite{Ref38}. This enables effective mitigation of spectral bias and facilitates frequency-aware operations~\cite{Ref14, Ref39}. Such designs demonstrate strong capability in recovering subtle textures and thin structures across diverse tasks~\cite{Ref15, Ref16}. However, their applicability is often hindered by rigid structural requirements, including fixed bases, tight-frame constraints, and high integration costs~\cite{Ref03, Ref17}. Consequently, while wavelet-based architectures excel at preserving high-frequency details, their constrained design limits their broader adoption as versatile decoders.

\subsubsection{Problem Statement.}
\label{sec2.4}
Building on the limitations outlined above, a structural mismatch emerges between practical decoder architectures and those designed to preserve high-frequency information. Practical architectures often compromise fine-detail fidelity, whereas high-fidelity architectures struggle with ease of integration. Bridging this gap calls for a decoder design that inherits the precision of wavelet methods while retaining the flexibility required for seamless integration into existing encoder--decoder pipelines.

\section{Proposed Method}
\label{sec:Proposed}

Classical wavelet transforms enforce Perfect Reconstruction (PR, $\Psi\Phi = I$), guaranteeing an identity mapping where the output perfectly matches the input. However, learning-based inverse problems require a non-identity mapping to recover a target signal $x$ from a degraded observation $y = Ax + \epsilon$. As established in the Deep Convolutional Framelets framework~\cite{Ref03}, successful deep networks must inherently violate the tight-frame condition to filter out task-irrelevant components. Strictly enforcing PR in U-Net decoders therefore creates a conflict between exact structural preservation and the necessity of feature suppression.

To resolve this, we propose \textbf{Selectively Suppressed Perfect Reconstruction (SUPER)}, an explicit formulation that relaxes PR as $\Psi\Phi = I - S$. In this framework, the wavelet filter bank ($I$) guarantees high-frequency integrity, while a learnable operator ($S$) selectively subtracts task-irrelevant features. This explicit decoupling of signal preservation and suppression enables the design of the \textbf{SUPER Module}, a plug-and-play block for U-Net variants. By replacing standard spatial upscaling with this wavelet-domain feature fusion process, the module maintains structural fidelity while improving overall computational efficiency.

\begin{figure}[tb]
  \centering
  \includegraphics[width=\linewidth]{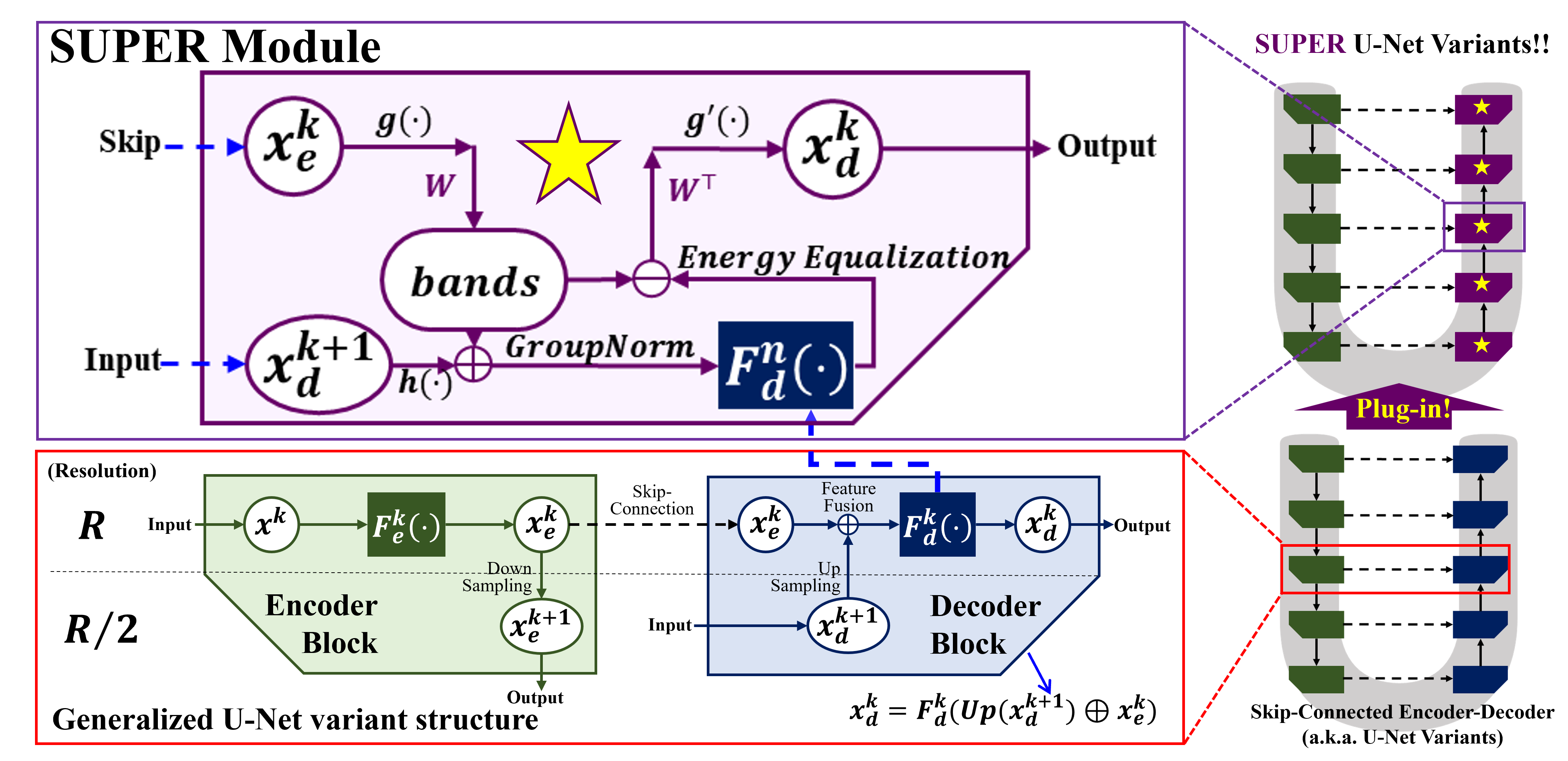}
  \caption{\textbf{Integration of the SUPER Module in generalized U-Net variants.} The proposed module directly replaces the standard decoder block in an encoder--decoder architecture, substituting spatial upscaling with wavelet-domain selective suppression to preserve structural integrity.}
  \label{fig:architecture}
\end{figure}

\subsection{Structural Formulation of SUPER}
\label{sec3.1}
To establish the mathematical foundation of SUPER, we first formalize the resolution recovery process of standard U-Net decoders. 
This formulation serves as the basis for analyzing the implicit violation of perfect reconstruction conditions during spatial upscaling.

\subsubsection{Generalization of U-Net Variants.}
Skip-connected architectures recover spatial resolution by fusing an encoder skip feature $x_e^k$ with an upscaled deeper feature $x_d^{k+1}$:
$$x_d^k = F_d^k(x_e^k \oplus U(x_d^{k+1})), $$
where $U(\cdot)$, $\oplus$, and $F_d^k(\cdot)$ represent the upscaling operator, feature fusion, and decoder transformation, respectively.

\subsubsection{Theoretical Relaxation of Tight Frames.}
Deep Convolutional Framelets (DCF) formally interpret encoder--decoder networks as paired frame operators $\Psi$ (decoder) and $\Phi$ (encoder). While classical wavelet theory dictates the tight-frame condition for exact PR ($\Psi \Phi = I$), the DCF framework demonstrates that effective inverse problem solving requires filtering operations that cannot equate to the identity matrix. Because the macro-level objective ($y = Ax + \epsilon$) necessitates discarding specific signal components, the intermediate operators must inherently deviate from the exact PR condition to isolate and remove task-irrelevant representations. However, unstructured implicit violations (e.g., pooling, non-linearities) lack explicit frequency-band control, inadvertently distorting crucial high-frequency details. To prevent this, we explicitly relax the PR constraint into a structured residual formulation: $\Psi \Phi = I - S$.

\subsubsection{Formulation of SUPER.}
For theoretical clarity, we first assume that the channel dimensions of the encoder skip feature ($C_{\text{e}}$) and the decoder output ($C_{\text{d}}$) are equal ($C_{\text{e}} = C_{\text{d}}$) to define the core PR relaxation mechanism. Here, the rigorous structure of the wavelet filter bank provides the exact preservation prior ($I$), while $S$ is explicitly introduced to isolate and subtract the task-irrelevant components.
At stage $k$, SUPER replaces conventional spatial upscaling with the inverse discrete wavelet transform (IDWT, denoted $W^\top$). To align spatial dimensions prior to fusion, the high-resolution skip feature $x_e^k$ is first decomposed by the DWT (denoted $W$).

In this formulation, the orthogonal wavelet operators map strictly to the exact preservation prior ($\Phi \equiv W, \Psi \equiv W^\top$), and the direct alias-aware fusion is formulated as:
$$x_d^k = W^\top\left(W(x_e^k) - F_d^k\left(W(x_e^k) \oplus x_d^{k+1}\right)\right),$$
where the IDWT operator ($W^\top$) rigorously guarantees alias-aware high-frequency preservation ($I$) during resolution recovery.

Crucially, $F_d^k$ operates directly on the decomposed frequency subbands as a structured residual modifier. 
This distinction is mathematically vital: any unsuppressed frequency component undergoes exact PR via the orthogonal wavelet filter bank in this idealized setting. Utilizing the deeper semantic feature $x_d^{k+1}$ as a global contextual guide, $F_d^k$ acts as a selective frequency suppressor ($S$), distinguishing between target structures to preserve and redundant nuisances to discard. 

Algebraically, one might initially compare our $\Psi\Phi = I - S$ formulation to standard spatial residual learning ($y = x + R$). 
However, this comparison differs critically due to the nature of the spatial expansion operator. 
Standard U-Net decoders lack an exact identity path during resolution recovery; their baseline is a spatially upsampled feature $U(x)$ that is inherently corrupted by aliasing and spectral bias. 
Consequently, a spatial residual $R$ must shoulder the ill-posed burden of simultaneously correcting interpolation artifacts and hallucinating lost high-frequency details. 

In stark contrast, SUPER operates within the orthogonal wavelet domain where the baseline state ($S=0$) mathematically guarantees Perfect Reconstruction ($I$) for the idealized $C_{\text{e}} = C_{\text{d}}$ case. 
The network is effectively relieved from the ill-posed task of synthesizing missing structures. 
Instead, it is tasked solely with bounded suppression—explicitly carving out task-irrelevant noise from a mathematically preserved high-frequency state. This shifts the decoder's fundamental objective from 'error correction' to 'selective filtration'.

\subsubsection{Amplitude-Bounded Stability.}
To prevent the network from aggressively over-subtracting informative features, we bound the suppression mechanism by applying a $\tanh$ activation to $F_d^k$. This restricts the module to act strictly as a bounded residual modifier rather than an unconstrained spatial distortion. Notably, when the network requires no suppression ($F_d^k(\cdot) = 0$), the operator defaults to $S=0$. In this theoretical identity state, the module guarantees that essential high-frequency spatial details are passed unaltered ($x_d^k = x_e^k$), ensuring structural preservation when deeper semantic modifications are unnecessary.

\subsubsection{Formal Stability Analysis.}
We define an analytical bound on reconstruction deviation within the projected subspace (\cref{sec3.2}), denoting the pre-expansion input and output as $\tilde{x}_e^k = g(x_e^k)$ and $\tilde{x}_d^k$. Orthonormal wavelets (e.g., Haar) preserve the $\ell_2$-norm via exact isometry ($\|W^\top z\|_2 = \|z\|_2$). Since the $\tanh$ activation strictly bounds the suppression tensor elements to $[-1, 1]$, the maximum deviation induced by $S$ is geometrically bounded by its volume and the maximum learnable scale $\|s_\lambda\|_\infty$:
$$\|\tilde{x}_d^k - \tilde{x}_e^k\|_2 \le \|s_\lambda\|_\infty \cdot \sqrt{C_{\text{sub}} \cdot H_{\text{sub}} \cdot W_{\text{sub}}}.$$
Although symmetric boundary padding introduces minor deviations from perfect isometry, the global end-to-end deviation remains constrained by the operator norms of $g$ and $g^\prime$. This formulation mathematically decouples the stability bound from the internal Lipschitz variations of $F_d^k$, ensuring stable feature refinement across stages.

\subsection{Plug-and-Play Modular Design}
\label{sec3.2}
While the formulation in \cref{sec3.1} defines the core frequency-domain identity path assuming $C_{\text{e}} = C_{\text{d}}$, practical U-Net variants require flexibility across diverse channel capacities. To address this, SUPER is implemented as an plug-and-play module for U-Net variants through adaptive bottlenecking and energy-calibration mechanisms. 
Its implementation consists of three operational stages: adaptive bottleneck compression, alias-free contextual fusion, and energy-calibrated suppression.

\subsubsection{Adaptive Bottleneck and Computational Efficiency.}
To align arbitrary channel capacities ($C_{\text{e}} \neq C_{\text{d}}$) efficiently, we introduce $1\times1$ bottleneck projections $g(\cdot)$ and $g'(\cdot)$. The projector $g(\cdot)$ compresses the encoder skip feature ($C_{\text{e}}$) into a $C_{\text{d}}/4$ subspace, which the DWT inherently expands back to $C_{\text{d}}$ across four subbands. Following reconstruction, $g^\prime(\cdot)$ restores the output to $C_{\text{d}}$. By processing features at half spatial resolution ($H/2 \times W/2$), the suppression operator $F_d^k$ reduces core MACs by $75\%$ (e.g., from $9C_{\text{d}}^2HW$ to $2.25C_{\text{d}}^2HW$ for $3\times3$ convolutions). 

Because the bottleneck projections $g$ and $g'$ perform channel compression, assuming them to act as exact inverses is mathematically ill-posed due to null-space information loss. Instead, $g$ and $g'$ are optimized to project features into an information-preserving subspace. Consequently, the module provides an approximate structural fallback to the baseline state from the perspective of the original feature space, and an exact identity within this projected subspace ($x_d^k = g'(g(x_e^k))$). This ensures that essential high-frequency spatial details are largely preserved when deeper semantic modifications are unnecessary. This spatial reduction guarantees net computational savings that scale with the complexity of $F_d^k$, while processing in the DWT domain intrinsically doubles the effective receptive field.

\subsubsection{Architecture-Agnostic Alias-Free Fusion.}
SUPER bypasses upscaling artifacts by seamlessly adapting to the host network's fusion paradigm without contaminating the exact PR identity path. To align the deeper semantic feature $x_d^{k+1}$ with the subband capacity, we introduce an auxiliary $1\times1$ projector $h(\cdot)$ that transforms $x_d^{k+1}$ into the $C_{out}/4$ subspace. 

For architectures utilizing \textit{concatenation}, this projected feature acts as a global contextual guide alongside the four wavelet subbands, resulting in an input tensor with $G=5$ components. For \textit{summation}-based architectures, $h(x_d^{k+1})$ is injected exclusively into the $LL$ subband ($G=4$). This implementation is mathematically significant: by isolating the deeper feature within the low-frequency approximation or as an auxiliary channel, the high-frequency bands ($LH, HL, HH$) remain untouched until the selective suppression stage. This design choice strictly preserves the PR fallback capability within the projected subspace, allowing the module to leverage deep semantic cues while maintaining alias-free structural fidelity.

\subsubsection{Subband Energy Equalization and Denormalization.}
Wavelet subbands exhibit extreme energy imbalance, with the $LL$ band typically dominating the sparse high-frequency details. To prevent the suppression operator $F_d^k$ from ignoring these subtle cues, we employ band-wise GroupNorm (GN) to equalize subband variances across the $G$ components. Following the refinement of the suppression mask by a CBAM module and applying the $\tanh$ bound, a learnable channel-wise scale vector $s_\lambda \in \mathbb{R}^{C_{out}}$ is applied. Crucially, these non-linear spatial and channel-wise transformations (GN, CBAM) are strictly confined to the formulation of the suppression mask ($S$). They do not directly alter the base subbands. Therefore, in the limiting case where $s_\lambda \to 0$ or $F_d^k \to 0$, the architecture mathematically defaults to the uncorrupted subbands, preserving the exact identity path within the projected subspace independent of the normalization dynamics.

\begin{algorithm}[tb]
    \caption{SUPER Module (stage $k$)}
    \label{alg:super_module}
    \textbf{Inputs:} $\mathbf{x}_e^k$ (encoder skip), $\mathbf{x}_d^{k+1}$ (decoder input) \\
    \textbf{Operator:} $F_d^k$ (generalized decoder block) \\
    \textbf{Parameters:} $g, h, g^\prime$ ($1\times1$ conv), $s_\lambda$ (learnable scale)
    \begin{algorithmic}[0] 
        \STATE \textbf{Compression:} $\mathbf{e}_{\text{comp}} \gets g(\mathbf{x}_e^k); \quad \mathbf{d}_{\text{comp}} \gets h(\mathbf{x}_d^{k+1})$ 
        
        \STATE \textbf{Decomposition:}  $\mathit{bands} \gets \text{DWT}(\mathbf{e}_{\text{comp}}) \quad \text{\COMMENT{$\mathit{bands} = [LL_e, LH_e, HL_e, HH_e]$}}$ 
        
        \STATE \textbf{Alias-Aware Fusion:}
        \IF{Fusion == Sum}
            \STATE $\mathbf{x}_{in} \gets \left[ LL_e + \mathbf{d}_{\text{comp}}, LH_e, HL_e, HH_e \right]; \quad G\gets4$
        \ELSIF{Fusion == Concat}
            \STATE $\mathbf{x}_{in} \gets \text{Concat}(\mathit{bands}, \mathbf{d}_{\text{comp}}); \quad G\gets5$
        \ENDIF
        
        \STATE \textbf{Energy Equalization:} $\mathbf{x}_{in} \gets \text{GroupNorm}(\mathbf{x}_{in}, \text{groups}=G)$
        
        \STATE \textbf{Selective Suppression:} $\mathit{sup} \gets \tanh\left(\text{CBAM}\left(F_d^k(\mathbf{x}_{in})\right)\right) \odot s_\lambda$ 
        
        \STATE \textbf{Reconstruction:} $\mathbf{out}_{\text{comp}} \gets \text{IDWT}(\mathit{bands} - \mathit{sup})$ 
        
        \STATE \textbf{Expansion:} $\mathbf{x}_d^k \gets g^\prime(\mathbf{out}_{\text{comp}})$
        \RETURN $\mathbf{x}_d^k$
    \end{algorithmic}
\end{algorithm}

\subsubsection{Implementation.}
The complete operational flow of SUPER Module is formulated as:
$$x_d^k = g^\prime\left( W^\top\left(W(g(x_e^k)) - \tanh\left(\mathcal{A}\left(F_d^k\left(\text{GN}\left(W(g(x_e^k)) \oplus h(x_d^{k+1})\right)\right)\right)\right) \odot s_\lambda \right) \right),$$
where $\mathcal{A}$ denotes the CBAM, $\text{GN}$ is band-wise GroupNorm, and $s_\lambda$ is the learnable scale vector. As detailed in \cref{fig:architecture} and \cref{alg:super_module}, decoupling mask generation from the reconstruction pathway enables the integration of non-invertible operations without compromising theoretical stability.

\section{Experimental Results}
\label{sec:Experiments}

Our primary objective is to validate SUPER as a \textbf{detail-sensitive}, \textbf{cost-efficient}, and consistently applicable \textbf{plug-and-play} decoder module.
We evaluate this across two core dimensions. First, we assess \textbf{detail enhancement and robustness} on three spectrally diverse tasks: monocular depth estimation (\cref{sec4.1}), extreme thin-crack segmentation (\cref{sec4.2}), and a low-frequency image denoising stress test (\cref{sec4.3}). 
Second, we demonstrate \textbf{cost-efficiency}, proving that bypassing spatial upscaling guarantees a highly favorable Pareto improvement by substantially reducing the target decoder's MACs and inference latency.

\subsubsection{Experimental Settings.}
For fair, structure-isolated comparisons, SUPER is integrated exclusively into the decoders of well-established baseline implementations under identical optimization settings. 
All experiments uniformly utilize the Haar wavelet, applying symmetric padding prior to the DWT to seamlessly handle boundary artifacts and odd-sized resolutions.
To accurately reflect our defined scope, computational costs (MACs and inference latency) are strictly standardized at a $512\times512\times3$ input resolution across all models. Latency is profiled on a single RTX 3090 GPU. We explicitly differentiate the target decoder's cost from the total model cost to isolate our architectural contribution. 
To ensure statistical reliability, results are averages of multiple runs, maintaining a relative standard deviation within 3\% for stability.
Extensive visualizations and the complete codebase will be made available on our project page.

\subsection{Monocular Depth Estimation (General Task)}
\label{sec4.1}
Monocular depth estimation requires a rigorous balance between global spatial context (for accurate scene geometry) and local boundary precision (for sharp object edges). This serves as an ideal general testbed to verify whether SUPER can actively enhance high-frequency details without compromising the low-frequency global structures of the host network.

\subsubsection{Dataset and Baselines.}
We evaluate our method on the widely used KITTI dataset, disentangling global and local performance by measuring general depth accuracy (A.Rel) on the standard Eigen split and fine-edge precision (Edge AUC) on the KITTI-DE benchmark. While we compare against established models (PackNet-SAN, AdaBins, PixelFormer) and their edge-enhanced variants, our primary baseline is \textbf{iDisc}~\cite{Ref29}. Although iDisc achieves exceptional global depth estimation by concentrating computation in its encoder, its reliance on standard spatial upscaling in the high-resolution decoder stages leaves it susceptible to edge degradation (as evidenced by its low baseline AUC of 26.59\%).

\subsubsection{Implementation.}
To isolate the architectural contribution, we incrementally replace the baseline's standard upscaling blocks with the SUPER Module components. 
The step-wise integration configurations sequentially integrate: (A) naive DWT, (B) Coupled Energy Calibration (GroupNorm + CBAM), (C) Selective Suppression, and (D) the full SUPER Module equipped with bottleneck projections.

\begin{table*}[t] 
    \centering
    \caption{\textbf{Quantitative results on the KITTI dataset for depth estimation.} General depth metrics are evaluated on the standard Eigen split of KITTI test set, and edge precision (AUC) is measured on the KITTI-DE split. (A)--(D) denote the incremental integration of components into the SUPER Module.}
    \label{Result1}
    \resizebox{\textwidth}{!}{ 
    \begin{tabular}{ll cc ccc cc cc}
        \toprule
        \multirow{2}{*}{\textbf{Baseline}} & \multirow{2}{*}{\textbf{Modification}} 
        & \multicolumn{2}{c}{\textbf{KITTI-test}} 
        & \multicolumn{3}{c}{\textbf{KITTI-DE} (Edge-AUC)} 
        & \multicolumn{2}{c}{\textbf{Decoder Cost}} 
        & \multicolumn{2}{c}{\textbf{Total Model Cost}} \\
        \cmidrule(lr){3-4} \cmidrule(lr){5-7} \cmidrule(lr){8-9} \cmidrule(lr){10-11}
        & & \textbf{A.Rel} $\downarrow$ & $\delta_{1.25} \uparrow$ & \textbf{AUC} $\uparrow$ 
        & A.Rel $\downarrow$ & $\delta_{1.25} \uparrow$  
        & \textbf{MACs} $\downarrow$ & \textbf{Latency} $\downarrow$
        & \textbf{MACs} $\downarrow$ & \textbf{Latency} $\downarrow$\\
        \midrule
        \multirow{5}{*}{\textbf{Packnet-SAN}} 
        & Original \cite{Ref45}           & 6.17\% & 95.39\% & 47.56\% & \textbf{3.45\%} & \textbf{98.66\%} &  &  &  &   \\
        & Boosting-Depth (O) \cite{Ref46} & 11.10\% & 86.41\% & 46.04\% & 9.32\% & 88.90\% &  &  &  &   \\
        & Boosting-Depth (K) \cite{Ref47} & 8.33\% & 91.99\% & 36.19\% & 7.24\% & 93.62\% &  &  &  &   \\
        & GradientFusion \cite{Ref48}    & 7.18\% & 94.17\% & 44.51\% & 5.93\% & 95.66\% &  &  &  &   \\
        & Edge-Loss \cite{Ref47}         & \textbf{6.50\%} & \textbf{95.90\%} & \textbf{61.87\%} & 3.61\% & 98.53\% &  &  &  &   \\
        \midrule
        \multirow{2}{*}{\textbf{AdaBins}} 
        & Original \cite{Ref49}          & 6.28\% & 95.85\% & 41.23\% & 3.14\% & 98.78\% &  &  &  &   \\
        & Edge-Loss \cite{Ref47}         & \textbf{6.21\%} & \textbf{95.87\%} & \textbf{53.47\%} & \textbf{3.11\%} & \textbf{98.79\%} &  &  &  &   \\
        \midrule
        \multirow{2}{*}{\textbf{PixelFormer}} 
        & Original \cite{Ref50}          & \textbf{5.45\%} & \textbf{96.98\%} & 32.79\% & 3.00\% & 98.79\% &  &  &  &   \\
        & Edge-Loss \cite{Ref47}         & 5.59\% & 96.72\% & \textbf{46.23\%} & \textbf{2.94\%} & \textbf{98.80\%} &  &  &  &   \\
        \midrule
        \multirow{5}{*}{\shortstack{\textbf{iDisc}}} 
        & Original \cite{Ref29}          & 5.83\% & 96.65\% & 26.59\% & 8.89\% & 98.62\% & 12.69 G & 17.02 ms & 344.04 G & 207.32 ms \\
        & \textbf{\textcolor{blue}{(A): DWT}}                         & 6.32\% & 95.89\% & 25.82\% & 9.21\% & 98.42\% & 52.25 G & 74.77 ms & 383.60 G & 282.57 ms \\
        & \textbf{\textcolor{blue}{(B): EE}}                    & 5.89\% & 96.22\% & 28.57\% & 9.27\% & 98.37\% & 52.25 G & 79.54 ms & 383.60 G & 283.48 ms \\
        & \textbf{\textcolor{blue}{(C): SS}}                          & 5.74\% & 96.78\% & 30.20\% & 7.39\% & 98.85\% & 52.25 G & 79.57 ms & 383.60 G & 283.99 ms \\
        & \textbf{\textcolor{blue}{(D): Bottleneck}}                  & \textbf{5.51\%} & \textbf{96.70\%} & \textbf{34.45\%} & \textbf{7.23\%} & \textbf{98.95\%} & \textbf{4.36 G} & \textbf{4.32 ms} & \textbf{335.7 G} & \textbf{198.72 ms} \\
        \bottomrule
    \end{tabular}
    }
\end{table*}

\subsubsection{Quantitative Results.}
Quantitative comparisons are summarized in \cref{Result1}. The fully equipped iDisc-(D) achieves the highest Edge AUC of 34.45\% on KITTI-DE, yielding a substantial absolute improvement of $+7.86\%$ over the original iDisc baseline. Concurrently, it improves the global A.Rel on KITTI-Test from 5.83\% to 5.51\%. Crucially, this simultaneous improvement in both global and local metrics is achieved while reducing the target decoder's MACs by 65.6\% (from 12.69\,G down to 4.36\,G).

\subsection{Thin-Crack Segmentation (Detail-Critical Task)}
\label{sec4.2}
Thin-crack segmentation provides a primary benchmark for evaluating high-frequency preservation. In this domain, minor upscaling artifacts and spatial distortions significantly degrade topological continuity. Therefore, it serves as an optimal task to verify the capability of the SUPER Module in maintaining fine structural fidelity.

\subsubsection{Dataset and Baselines.}
We evaluate our method on the CrackVision12K dataset, prioritizing the challenging sub-4\,px categories (0--2\,px and 2--4\,px). To ensure a comprehensive assessment of structural fidelity, we employ both IoU and the Crack Topology Score (CTS). As established in centerline-aware segmentation studies~\cite{shit2021cldice}, these metrics exhibit inherent scale biases: IoU is highly sensitive to minor pixel shifts in thin structures, while centerline-based metrics like CTS can become unreliable for thick cracks where boundary irregularities trigger spurious skeleton branches. 

Our primary baseline is \textbf{FACS-Net}~\cite{Ref24}, a frequency-aware architecture optimized for fine crack recovery. To evaluate upscaling efficiency, we also benchmark against advanced dynamic spatial upscalers, CARAFE~\cite{CARAFE} and FADE~\cite{FADE}.

\subsubsection{Implementation.}
To isolate SUPER's architectural impact, we formulate \textbf{FACS-SUPER} by replacing the baseline decoder's Double Convolution blocks and attention layers with SUPER blocks, keeping the hybrid encoder unchanged. 
Critically, SUPER maintains a parameter count comparable to or lower than the original blocks to ensure gains stem from our frequency-domain formulation rather than expanded model capacity. 
All models are evaluated under identical training and optimization settings for a fair structural comparison.

\begin{table}[tb]
    \centering
    \caption{\textbf{Quantitative results and hardware profiling on CrackVision12K dataset.} In addition to IoU and Crack Topology Score (CTS), we report the empirical hardware metrics.}
    \label{Result2}
    \resizebox{\textwidth}{!}{  
    \setlength{\tabcolsep}{12pt}
    \begin{tabular}{cl ccc cc cc}
        \toprule
        \multirow{2}{*}{\textbf{Method}} & \multirow{2}{*}{\textbf{Metric}} 
        & \multicolumn{3}{c}{\textbf{Crack Width (px)}} 
        & \multicolumn{2}{c}{\textbf{Decoder Cost}} 
        & \multicolumn{2}{c}{\textbf{Total Model Cost}}  \\
        \cmidrule(lr){3-5} \cmidrule(lr){6-7} \cmidrule(lr){8-9}
        & & \textbf{0--2} & \textbf{2--4} & \textbf{Average} 
        & \textbf{MACs $\downarrow$} & \textbf{Latency $\downarrow$} & \textbf{MACs $\downarrow$} & \textbf{Latency $\downarrow$} \\
        \midrule
        \multirow{2}{*}{\textbf{\textcolor{blue}{FACS-SUPER}}} 
        & \textbf{IoU}  $\uparrow$ & \textbf{0.451} & \textbf{0.415} & \textbf{0.646} & \multirow{2}{*}{53.65 G} & \multirow{2}{*}{\textbf{28.42 ms}} & \multirow{2}{*}{484.15 G} & \multirow{2}{*}{\textbf{250.75 ms}} \\
        & \textbf{CTS} $\uparrow$ & \textbf{0.916} & \textbf{0.806} & 0.593 & & & & \\
        \addlinespace
        \multirow{2}{*}{FACS-FADE \cite{FADE, Ref24}} 
        & \textbf{IoU} $\uparrow$ & 0.397 & 0.401 & 0.628 & \multirow{2}{*}{\textbf{15.45 G}} & \multirow{2}{*}{56.52 ms} & \multirow{2}{*}{\textbf{445.95 G}} & \multirow{2}{*}{280.92 ms}  \\
        & \textbf{CTS} $\uparrow$ & 0.901 & 0.798 & \textbf{0.614} & & & & \\
        \addlinespace
        \multirow{2}{*}{FACS-CARAFE \cite{CARAFE, Ref24}} 
        & \textbf{IoU} $\uparrow$ & 0.373 & 0.403 & 0.629 & \multirow{2}{*}{406.72 G} & \multirow{2}{*}{108.28 ms} & \multirow{2}{*}{837.22 G} & \multirow{2}{*}{335.49 ms} \\
        & \textbf{CTS} $\uparrow$ & 0.899 & 0.799 & 0.557 & & & & \\
        \addlinespace
        \multirow{2}{*}{FACS-Net \cite{Ref24}} 
        & \textbf{IoU} $\uparrow$ & 0.356 & 0.398 & 0.629 & \multirow{2}{*}{156.56 G} & \multirow{2}{*}{29.23 ms} & \multirow{2}{*}{587.06 G} & \multirow{2}{*}{258.18 ms} \\
        & \textbf{CTS} $\uparrow$ & 0.872 & 0.778 & 0.532 & & & & \\
        \midrule
        \addlinespace
        \multirow{2}{*}{Wavelet U-Net++ \cite{Ref16}} 
        & \textbf{IoU} $\uparrow$ & 0.382 & 0.417 & 0.608 & \multirow{2}{*}{ } & \multirow{2}{*}{ } & \multirow{2}{*}{ } & \multirow{2}{*}{ } \\
        & \textbf{CTS} $\uparrow$ & 0.912 & 0.793 & 0.630 & & & & \\
        \addlinespace
        \multirow{2}{*}{MWCNN \cite{Ref06}} 
        & \textbf{IoU} $\uparrow$ & 0.348 & 0.401 & 0.613 & \multirow{2}{*}{} & \multirow{2}{*}{} & \multirow{2}{*}{ } & \multirow{2}{*}{ } \\
        & \textbf{CTS} $\uparrow$ & 0.899 & 0.754 & 0.625 & & & & \\
        \addlinespace
        \multirow{2}{*}{DECS-Net \cite{Ref55}} 
        & \textbf{IoU} $\uparrow$ & 0.275 & 0.301 & 0.564 & \multirow{2}{*}{ } & \multirow{2}{*}{ } & \multirow{2}{*}{ } & \multirow{2}{*}{ } \\
        & \textbf{CTS} $\uparrow$ & 0.896 & 0.809 & 0.626 & & & & \\
        \midrule
        \addlinespace
        \multirow{2}{*}{\shortstack{Hybrid-Segmentor \\ \cite{Ref51} (BCE+Dice)}} 
        & \textbf{IoU} $\uparrow$ & 0.160 & 0.365 & 0.625 & \multirow{2}{*}{ } & \multirow{2}{*}{ } & \multirow{2}{*}{ } & \multirow{2}{*}{ } \\
        & \textbf{CTS} $\uparrow$ & 0.585 & 0.819 & 0.619 & & & & \\
        \addlinespace
        \multirow{2}{*}{DeepLabV3+ \cite{Ref54}} 
        & \textbf{IoU} $\uparrow$ & 0.079 & 0.301 & 0.595 & \multirow{2}{*}{ } & \multirow{2}{*}{ } & \multirow{2}{*}{ } & \multirow{2}{*}{ } \\
        & \textbf{CTS} $\uparrow$ & 0.101 & 0.809 & 0.490 & & & & \\
        \addlinespace
        \bottomrule
    \end{tabular}
    }
\end{table}

\subsubsection{Quantitative Results.}
Quantitative comparisons and hardware profiling are detailed in \cref{Result2}. FACS-SUPER outperforms all evaluated benchmarks in average IoU (0.646), demonstrating superior overall structural accuracy. Notably, in the most extreme 0--2\,px regime, FACS-SUPER achieves a strong CTS of 0.916, maintaining a clear advantage over both traditional wavelet architectures (Wavelet U-Net++, MWCNN; CTS of 0.912 and 0.899) and standard non-frequency-aware models. 

Regarding hardware efficiency, FACS-SUPER records the lowest overall and decoder latency. While FACS-FADE maintains lower decoder MACs, FACS-SUPER achieves a substantial 65.7\% reduction in decoder MACs compared to the original FACS-Net while simultaneously delivering significantly higher structural accuracy and faster inference.

\subsection{Smartphone Image Denoising (Stress-Test)}
\label{sec4.3}
Smartphone image denoising serves as a rigorous low-frequency stress test. Because detail-enhancing modules often overemphasize high-frequency features and amplify noise in smooth regions, this task evaluates SUPER's robustness and structural stability when high-frequency cues are inherently scarce.

\subsubsection{Dataset and Baselines.}
We evaluate on the Smartphone Image Denoising Dataset (SIDD) using standard Peak Signal-to-Noise Ratio (PSNR) and Structural Similarity Index (SSIM) metrics. 
Crucially, the ground-truth images in SIDD are generated via multi-frame averaging, which naturally smooths out fine high-frequency details. Under this low-frequency-dominant regime, we benchmark against recent state-of-the-art architectures (Restormer, NAFNet). Our primary baseline is \textbf{CascadedGaze-Net (CG-Net)}~\cite{Ref31}, a highly competitive, fully CNN-based model explicitly optimized for aggressive noise removal and global luminance coherence.

\subsubsection{Implementation.}
CG-Net consists of an encoder based on GCE blocks and a decoder based on NAF blocks~\cite{Ref62}. To construct \textbf{CG-SUPER}, we strictly retain the original encoder and replace the standard NAF blocks in each decoder stage with our SUPER blocks, utilizing the NAF block structure as the internal suppression operator $F_d^k$.

\begin{table}[tb]
    \centering
    \caption{\textbf{Quantitative results on SIDD (image denoising).} Comparison of CG-SUPER with recent state-of-the-art denoising models.}
    \label{Result3}
    \resizebox{\textwidth}{!}{ 
    \setlength{\tabcolsep}{12pt}
    \begin{tabular}{l cc cc cc}
        \toprule
        \multirow{2}{*}{\textbf{Method}} 
        & \multicolumn{2}{c}{\textbf{Quality Metrics}} 
        & \multicolumn{2}{c}{\textbf{Decoder Cost}}
        & \multicolumn{2}{c}{\textbf{Total Cost}} \\
        \cmidrule(lr){2-3} \cmidrule(lr){4-5} \cmidrule(lr){6-7}
        & \textbf{PSNR} $\uparrow$ & \textbf{SSIM} $\uparrow$ & \textbf{MACs $\downarrow$} & \textbf{Latency $\downarrow$} & \textbf{MACs $\downarrow$} & \textbf{Latency $\downarrow$}  \\
        \midrule
        \textbf{\textcolor{blue}{CG-SUPER}} & \begin{tabular}{@{}c@{}} \textbf{40.416}\\ \scriptsize{$\pm$ 0.003 (n=10)} \end{tabular} & 0.964 & \textbf{9.35 G} & \textbf{11.46 ms} & \textbf{223.04 G} & \textbf{107.11 ms} \\
        \addlinespace
        CascadedGaze \cite{Ref31} & 40.39 & 0.964 & 28.03 G & 19.02 ms & 249.26 G & 124.07 ms \\
        \midrule
        NAFNet \cite{Ref62}       & 40.30 & 0.962 &  &  &  &  \\
        Restormer \cite{Ref63}    & 40.02 & 0.960 &  &  &  &  \\
        \bottomrule
    \end{tabular}
    }
\end{table}

\subsubsection{Quantitative Results.}
Quantitative comparisons are detailed in \cref{Result3}. Despite the highly challenging, low-frequency-dominant environment, CG-SUPER avoids any performance degradation, achieving a PSNR of 40.416\,dB and an SSIM of 0.964, slightly outperforming the baseline CG-Net (40.39\,dB). 
More importantly, this structural stability is achieved while massively reducing the decoder's computational cost: CG-SUPER operates at merely 9.35\,G MACs, representing a substantial 66.64\% reduction compared to the baseline decoder's 28.03\,G MACs.

\section{Discussion}
\label{sec:Discussion}
The empirical results in \cref{sec:Experiments} confirm that the \textbf{plug-and-play} integration of SUPER consistently delivers a highly favorable \textbf{Pareto improvement} across diverse dense prediction tasks and completely different host architectures (iDisc, FACS-Net, and CG-Net). In this section, we analyze the underlying structural mechanisms that enable these gains and discuss the robustness and theoretical limits of our formulation.

\subsection{Generalization and Detail-Cost Trade-off}
\label{sec5.1}

\subsubsection{Monocular Depth Estimation Mechanics.}
Monocular depth estimation effectively exposes the inherent flaw of conventional upscaling. As shown in \cref{Result1}, the baseline iDisc achieves highly competitive global depth accuracy (A.Rel) but records the lowest fine-edge precision (AUC), illustrating the detail-blurring limitation of standard U-Net decoders. 

SUPER dismantles this traditional accuracy-efficiency trade-off. By replacing upscaling blocks, it significantly elevates Edge AUC while simultaneously reducing the target decoder's MACs and inference latency. Crucially, global A.Rel inherently improves as well, proving SUPER injects fine-detail sensitivity without compromising global semantic capacity.

\subsubsection{Incremental Study Analysis.}
The step-wise integration uncovers the necessity of our architectural formulations. Replacing spatial upscaling with naive DWT (A) degrades metrics due to uncalibrated subband energy imbalances. Coupled Energy Calibration (B)—pairing GroupNorm with CBAM—rectifies this discrepancy and stabilizes representations. Selective Suppression (C) then isolates task-relevant high frequencies, substantially boosting edge recovery. Finally, the bottleneck (D) preserves these enhancements while structurally reducing the computational footprint. This validates SUPER as a rigorously controlled, frequency-aware design rather than a heuristic coincidence.

\subsection{Fine-Detail Sensitivity Beyond Upscaling Limits}
\label{sec5.2}

\subsubsection{Thin-Crack Segmentation Analysis.}
As shown in \cref{Result2}, FACS-SUPER achieves the highest average IoU (0.646) and demonstrates superior precision in the critical 0--2\,px regime. While traditional wavelet models (e.g., MWCNN) report competitive CTS, these scores are often inflated by centerline artifacts on thick cracks, as reflected in their lower overall IoU. In contrast, FACS-SUPER provides a more robust balance between local topological continuity and global structural consistency.

From a structural perspective, SUPER avoids the symmetric encoder--decoder constraints of traditional wavelet architectures, enabling seamless integration with heterogeneous backbones. Furthermore, while dynamic upscalers (CARAFE, FADE) suffer from spatial-domain latency, SUPER's decoder-centric wavelet restoration bypasses interpolation artifacts while maintaining high computational efficiency.

\subsection{Low-Frequency Robustness and Efficiency}
\label{sec5.3}

\subsubsection{Smartphone Image Denoising Analysis.}
Smartphone image denoising (SIDD) is a highly challenging stress test: its multi-frame averaged ground truths lack high-frequency details, making detail enhancement unachievable and rendering residual noise highly prominent.
However, SUPER acts as an adaptive safe filter. Instead of suffering severe PSNR degradation typical of overfitted models, CG-SUPER maintains global stability and marginally improves performance. This proves our bounded suppression explicitly recognizes the absence of target structures and safely scales back.

Crucially, even when dataset characteristics limit further structural gains, SUPER achieves a 66.6\% reduction in decoder MACs and improves inference speed. This demonstrates that the module can yield a consistent Pareto improvement, providing significant computational relief while maintaining robust performance across diverse task regimes.

\subsection{Strengths, Limitations, and Practical Value}
\label{sec5.4}

SUPER offers a \textbf{plug-and-play} solution for \textbf{detail-sensitive} recovery and \textbf{cost-efficient} inference. Across diverse tasks, it yields a consistent \textbf{Pareto improvement}, balancing fine-detail fidelity with significantly reduced computational overhead.

The module is inherently robust: in scenarios where data limits further structural refinement, SUPER preserves global context without introducing artifacts. Moreover, while its MAC reductions are localized to the decoder, they provide guaranteed and predictable efficiency gains within the target stages. Ultimately, SUPER provides a versatile mechanism that enhances local precision when supported by data while ensuring consistent computational savings, establishing it as a robust structural alternative for modern dense prediction.

\section{Conclusion}
\label{sec:Conclusion}
This paper addresses the limitations of fine-structure recovery caused by spatial upscaling in U-Net variants by proposing the Selectively Suppressed Perfect Reconstruction (SUPER) framework. By relaxing the strict perfect reconstruction (PR) condition of classical wavelet transforms into a learnable suppression operator, we establish a mathematical foundation that explicitly disentangles structural signal preservation from the suppression of task-irrelevant features.

Crucially, the proposed SUPER module features an effective integration design that can readily replace the decoder blocks of widely used U-Net variants without requiring comprehensive architectural redesigns. Experimental results across diverse dense prediction tasks confirm that this module seamlessly integrates into heterogeneous host networks, achieving a consistent Pareto improvement by substantially reducing decoder computational costs while simultaneously enhancing fine-detail recovery.

While SUPER effectively recovers available high-frequency information, its capacity to hallucinate structures entirely absent from the encoder inputs remains bounded due to its reliance on skip connections. Future work will explore extending this framework with generative frequency priors or diverse wavelet bases (e.g., biorthogonal framelets) to further broaden its technical applicability.

%
%
\bibliographystyle{splncs04}
\bibliography{main}

@String(CVPR  = {IEEE Conf. Comput. Vis. Pattern Recog.})

@String(ICCV  = {Int. Conf. Comput. Vis.})

@String(ECCV  = {Eur. Conf. Comput. Vis.})

@String(ICIP  = {IEEE Int. Conf. Image Process.})

@String(CVPR  = {CVPR})

@String(ICCV  = {ICCV})

@String(ECCV  = {ECCV})

@String(ICIP  = {ICIP})

@article{Ref01,
  author = {A. A. Baniya and T.-K. Lee and P. Eklund and S. Aryal},
  title = {A Survey of Deep Learning Video Super-Resolution},
  journal = {IEEE Transactions on Emerging Topics in Computational Intelligence},
  volume = {8},
  number = {4},
  pages = {2655--2676},
  month = {Aug},
  year = {2024},
  doi = {10.1109/TETCI.2024.3398015}
}

@article{Ref02,
  author = {A. Dede and others},
  title = {Deep learning for efficient high-resolution image processing: A systematic review},
  journal = {Intelligent Systems with Applications},
  volume = {26},
  pages = {200505},
  month = {Jun},
  year = {2025},
  doi = {10.1016/j.iswa.2025.200505}
}

@article{Ref03,
  author = {J. C. Ye and Y. Han and E. Cha},
  title = {Deep Convolutional Framelets: A General Deep Learning Framework for Inverse Problems},
  journal = {SIAM Journal on Imaging Sciences},
  volume = {11},
  number = {2},
  pages = {991--1048},
  month = {Jan},
  year = {2018},
  doi = {10.1137/17M1141771}
}

@article{Ref04,
  author = {O. Ronneberger and P. Fischer and T. Brox},
  title = {U-Net: Convolutional Networks for Biomedical Image Segmentation},
  journal = {arXiv preprint arXiv:1505.04597},
  month = {May},
  year = {2015},
  doi = {10.48550/arXiv.1505.04597}
}

@article{Ref05,
  author = {W. Jiangtao and N. I. R. Ruhaiyem and F. Panpan},
  title = {A Comprehensive Review of U-Net and Its Variants: Advances and Applications in Medical Image Segmentation},
  journal = {IET Image Processing},
  volume = {19},
  number = {1},
  pages = {e70019},
  year = {2025},
  doi = {10.1049/ipr2.70019}
}

@article{Ref06,
  author = {P. Liu and H. Zhang and K. Zhang and L. Lin and W. Zuo},
  title = {Multi-level Wavelet-CNN for Image Restoration},
  journal = {arXiv preprint arXiv:1805.07071},
  month = {May},
  year = {2018},
  doi = {10.48550/arXiv.1805.07071}
}

@misc{Ref07,
  author = {F. Zhang and S. B. Rangrej and T. Aumentado-Armstrong and A. Fazly and A. Levinshtein},
  title = {Augmenting Perceptual Super-Resolution via Image Quality Predictors}
}

@article{Ref08,
  author = {C. Zhang and X. Deng and S. H. Ling},
  title = {Next-Gen Medical Imaging: U-Net Evolution and the Rise of Transformers},
  journal = {Sensors},
  volume = {24},
  number = {14},
  pages = {4668},
  month = {Jul},
  year = {2024},
  doi = {10.3390/s24144668}
}

@inproceedings{Ref09,
  author = {X. Lin and Y. Li and J. Hsiao and C. Ho and Y. Kong},
  title = {Catch Missing Details: Image Reconstruction with Frequency Augmented Variational Autoencoder},
  booktitle = {2023 IEEE/CVF Conference on Computer Vision and Pattern Recognition (CVPR)},
  address = {Vancouver, BC, Canada},
  publisher = {IEEE},
  month = {Jun},
  year = {2023},
  pages = {1736--1745},
  doi = {10.1109/CVPR52729.2023.00173}
}

@inproceedings{Ref10,
  author = {N. Rahaman and others},
  title = {On the Spectral Bias of Neural Networks},
  booktitle = {Proceedings of the 36th International Conference on Machine Learning},
  publisher = {PMLR},
  month = {May},
  year = {2019},
  pages = {5301--5310},
  url = {https://proceedings.mlr.press/v97/rahaman19a.html}
}

@article{Ref11,
  author = {Y. Cao and Z. Fang and Y. Wu and D.-X. Zhou and Q. Gu},
  title = {Towards Understanding the Spectral Bias of Deep Learning},
  journal = {arXiv preprint arXiv:1912.01198},
  month = {Oct},
  year = {2020},
  doi = {10.48550/arXiv.1912.01198}
}

@inproceedings{Ref12,
  author = {H.-H. Yang and Y. Fu},
  title = {Wavelet U-Net and the Chromatic Adaptation Transform for Single Image Dehazing},
  booktitle = {2019 IEEE International Conference on Image Processing (ICIP)},
  month = {Sep},
  year = {2019},
  pages = {2736--2740},
  doi = {10.1109/ICIP.2019.8803391}
}

@inproceedings{Ref13,
  author = {X. Deng and R. Yang and M. Xu and P. L. Dragotti},
  title = {Wavelet Domain Style Transfer for an Effective Perception-Distortion Tradeoff in Single Image Super-Resolution},
  booktitle = {2019 IEEE/CVF International Conference on Computer Vision (ICCV)},
  address = {Seoul, Korea (South)},
  publisher = {IEEE},
  month = {Oct},
  year = {2019},
  pages = {3076--3085},
  doi = {10.1109/ICCV.2019.00317}
}

@misc{Ref14,
  author = {R. Fang and Y. Xu},
  title = {Addressing Spectral Bias of Deep Neural Networks by Multi-Grade Deep Learning}
}

@article{Ref15,
  author = {Y. Zhao and S. Wang and Y. Zhang and S. Qiao and M. Zhang},
  title = {WRANet: wavelet integrated residual attention U-Net network for medical image segmentation},
  journal = {Complex \& Intelligent Systems},
  volume = {9},
  number = {6},
  pages = {6971--6983},
  month = {Dec},
  year = {2023},
  doi = {10.1007/s40747-023-01119-y}
}

@article{Ref16,
  author = {S. Akila Agnes and A. Arun Solomon and K. Karthick},
  title = {Wavelet U-Net++ for accurate lung nodule segmentation in CT scans: Improving early detection and diagnosis of lung cancer},
  journal = {Biomedical Signal Processing and Control},
  volume = {87},
  pages = {105509},
  month = {Jan},
  year = {2024},
  doi = {10.1016/j.bspc.2023.105509}
}

@inproceedings{Ref17,
  author = {E. Marcus and R. Sheombarsing and J.-J. Sonke and J. Teuwen},
  title = {Task-Driven Wavelets Using Constrained Empirical Risk Minimization},
  booktitle = {2024 IEEE/CVF Conference on Computer Vision and Pattern Recognition (CVPR)},
  address = {Seattle, WA, USA},
  publisher = {IEEE},
  month = {Jun},
  year = {2024},
  pages = {24098--24107},
  doi = {10.1109/CVPR52733.2024.02275}
}

@inproceedings{Ref18,
  author = {L. Jiang and B. Dai and W. Wu and C. C. Loy},
  title = {Focal Frequency Loss for Image Reconstruction and Synthesis},
  booktitle = {2021 IEEE/CVF International Conference on Computer Vision (ICCV)},
  address = {Montreal, QC, Canada},
  publisher = {IEEE},
  month = {Oct},
  year = {2021},
  pages = {13899--13909},
  doi = {10.1109/ICCV48922.2021.01366}
}

@article{Ref19,
  author = {W. Yang and others},
  title = {Optimizing transformer-based network via advanced decoder design for medical image segmentation},
  journal = {Biomedical Physics \& Engineering Express},
  volume = {11},
  number = {2},
  pages = {025024},
  month = {Mar},
  year = {2025},
  doi = {10.1088/2057-1976/adaec7}
}

@article{Ref20,
  author = {J. C. Santos and H. Tomás Pereira Alexandre and M. Seoane Santos and P. Henriques Abreu},
  title = {The Role of Deep Learning in Medical Image Inpainting: A Systematic Review},
  journal = {ACM Transactions on Computing for Healthcare},
  volume = {6},
  number = {3},
  pages = {1--24},
  month = {Jul},
  year = {2025},
  doi = {10.1145/3712710}
}

@article{Ref21,
  author = {B. Yu and Q. Zhou and L. Yuan and H. Liang and P. Shcherbakov and X. Zhang},
  title = {3D medical image segmentation using the serial–parallel convolutional neural network and transformer based on cross-window self-attention},
  journal = {CAAI Transactions on Intelligence Technology},
  volume = {10},
  number = {2},
  pages = {337--348},
  month = {Apr},
  year = {2025},
  doi = {10.1049/cit2.12411}
}

@article{Ref22,
  author = {Z.-Q. J. Xu and L. Zhang and W. Cai},
  title = {On understanding and overcoming spectral biases of deep neural network learning methods for solving PDEs},
  journal = {arXiv preprint arXiv:2501.09987},
  month = {Jan},
  year = {2025},
  doi = {10.48550/arXiv.2501.09987}
}

@article{Ref23,
  author = {Y. Wu and S. Li and J. Li and Y. Yu and J. Li and Y. Li},
  title = {Deep learning in crack detection: A comprehensive scientometric review},
  journal = {Journal of Infrastructure Intelligence and Resilience},
  volume = {4},
  number = {3},
  pages = {100144},
  month = {Sep},
  year = {2025},
  doi = {10.1016/j.iintel.2025.100144}
}

@article{Ref24,
  author = {S. Joo and S. Kim and H. Kim},
  title = {Frequency-aware crack segmentation network (FACS-net) and crack topology loss (CT-loss) for thin cracks},
  journal = {Autom. Constr.},
  volume = {182},
  pages = {106719},
  month = {Feb},
  year = {2026},
  doi = {10.1016/j.autcon.2025.106719}
}

@article{Ref25,
  author = {A. Zaman and others},
  title = {Adaptive Feature Medical Segmentation Network: an adaptable deep learning paradigm for high-performance 3D brain lesion segmentation in medical imaging},
  journal = {Frontiers in Neuroscience},
  volume = {18},
  pages = {1363930},
  month = {Apr},
  year = {2024},
  doi = {10.3389/fnins.2024.1363930}
}

@inproceedings{Ref26,
  author = {L. Talker and A. Cohen and E. Yosef and A. Dana and M. Dinerstein},
  title = {Mind The Edge: Refining Depth Edges in Sparsely-Supervised Monocular Depth Estimation},
  booktitle = {2024 IEEE/CVF Conference on Computer Vision and Pattern Recognition (CVPR)},
  address = {Seattle, WA, USA},
  publisher = {IEEE},
  month = {Jun},
  year = {2024},
  pages = {10606--10616},
  doi = {10.1109/CVPR52733.2024.01009}
}

@article{Ref27,
  author = {J. Peng and M. Lu and B. Li and J. Wang and W. Hu and X. Liu},
  title = {Frequency-aware denoising using a diffusion model for enhanced band-limited and white noise removal in x-ray acoustic computed tomography},
  journal = {Medical Physics},
  volume = {52},
  number = {5},
  pages = {3325--3335},
  month = {May},
  year = {2025},
  doi = {10.1002/mp.17681}
}

@article{Ref29,
  author = {L. Piccinelli and C. Sakaridis and F. Yu},
  title = {iDisc: Internal Discretization for Monocular Depth Estimation},
  journal = {arXiv preprint arXiv:2304.06334},
  month = {Apr},
  year = {2023},
  doi = {10.48550/arXiv.2304.06334}
}

@article{Ref31,
  author = {A. Ghasemabadi and M. K. Janjua and M. Salameh and C. Zhou and F. Sun and D. Niu},
  title = {CascadedGaze: Efficiency in Global Context Extraction for Image Restoration},
  journal = {arXiv preprint arXiv:2401.15235},
  month = {May},
  year = {2024},
  doi = {10.48550/arXiv.2401.15235}
}

@article{Ref32,
  author = {O. T. Paalvast and O. Hertgers and M. Sevenster and H. J. Lamb},
  title = {Assessing the Image Quality of Digitally Reconstructed Radiographs from Chest CT},
  journal = {Journal of Imaging Informatics in Medicine},
  month = {Jan},
  year = {2025},
  doi = {10.1007/s10278-025-01406-9}
}

@article{Ref33,
  author = {I. Aganj and B. T. T. Yeo and M. R. Sabuncu and B. Fischl},
  title = {On Removing Interpolation and Resampling Artifacts in Rigid Image Registration},
  journal = {IEEE Transactions on Image Processing},
  volume = {22},
  number = {2},
  pages = {816--827},
  month = {Feb},
  year = {2013},
  doi = {10.1109/TIP.2012.2224356}
}

@article{Ref34,
  author = {A. H. Ribeiro and T. B. Sch{\"o}n},
  title = {How Convolutional Neural Networks Deal with Aliasing},
  journal = {arXiv preprint arXiv:2102.07757},
  month = {Feb},
  year = {2021},
  doi = {10.48550/arXiv.2102.07757}
}

@article{Ref35,
  author = {C. Tan and others},
  title = {Rethinking the Up-Sampling Operations in CNN-based Generative Network for Generalizable Deepfake Detection},
  journal = {arXiv preprint arXiv:2312.10461},
  month = {Dec},
  year = {2023},
  doi = {10.48550/arXiv.2312.10461}
}

@article{Ref36,
  author = {X. Gou and C. Liao and J. Zhou and F. Ye and Y. Lin},
  title = {FIF-UNet: An Efficient UNet Using Feature Interaction and Fusion for Medical Image Segmentation},
  journal = {arXiv preprint arXiv:2409.05324},
  month = {Sep},
  year = {2024},
  doi = {10.48550/arXiv.2409.05324}
}

@inproceedings{Ref37,
  author = {S. Sawant and A. Medgyesy and S. Raghunandan and T. G{\"o}tz},
  title = {LMSC-UNet: A Lightweight U-Net with Modified Skip Connections for Semantic Segmentation},
  booktitle = {Proceedings of the 17th International Conference on Agents and Artificial Intelligence},
  address = {Porto, Portugal},
  publisher = {SCITEPRESS - Science and Technology Publications},
  year = {2025},
  pages = {726--734},
  doi = {10.5220/0013343800003890}
}

@misc{Ref38,
  author = {J. Zeng and L. Huang and K. Wang},
  title = {WST: Wavelet-Based Multi-scale Tuning for Visual Transfer Learning}
}

@inproceedings{Ref39,
  author = {Q. Bu and D. Huang and H. Cui},
  title = {Towards Building More Robust Models with Frequency Bias},
  booktitle = {2023 IEEE/CVF International Conference on Computer Vision (ICCV)},
  address = {Paris, France},
  publisher = {IEEE},
  month = {Oct},
  year = {2023},
  pages = {4379--4388},
  doi = {10.1109/ICCV51070.2023.00406}
}

@inproceedings{Ref45,
  author = {V. Guizilini and R. Ambrus and S. Pillai and A. Raventos and A. Gaidon},
  title = {3D Packing for Self-Supervised Monocular Depth Estimation},
  booktitle = {2020 IEEE/CVF Conference on Computer Vision and Pattern Recognition (CVPR)},
  address = {Seattle, WA, USA},
  publisher = {IEEE},
  month = {Jun},
  year = {2020},
  pages = {2482--2491},
  doi = {10.1109/CVPR42600.2020.00256}
}

@inproceedings{Ref46,
  author = {S. M. H. Miangoleh and S. Dille and L. Mai and S. Paris and Y. Aksoy},
  title = {Boosting Monocular Depth Estimation Models to High-Resolution via Content-Adaptive Multi-Resolution Merging},
  booktitle = {2021 IEEE/CVF Conference on Computer Vision and Pattern Recognition (CVPR)},
  month = {Jun},
  year = {2021},
  pages = {9680--9689},
  doi = {10.1109/CVPR46437.2021.00956}
}

@inproceedings{Ref47,
  author = {L. Talker and A. Cohen and E. Yosef and A. Dana and M. Dinerstein},
  title = {Mind The Edge: Refining Depth Edges in Sparsely-Supervised Monocular Depth Estimation},
  booktitle = {2024 IEEE/CVF Conference on Computer Vision and Pattern Recognition (CVPR)},
  address = {Seattle, WA, USA},
  publisher = {IEEE},
  month = {Jun},
  year = {2024},
  pages = {10606--10616},
  doi = {10.1109/CVPR52733.2024.01009}
}

@article{Ref48,
  author = {Y. Dai and R. Yi and C. Zhu and H. He and K. Xu},
  title = {Multi-resolution Monocular Depth Map Fusion by Self-supervised Gradient-based Composition},
  journal = {arXiv preprint arXiv:2212.01538},
  year = {2022},
  doi = {10.48550/ARXIV.2212.01538}
}

@inproceedings{Ref49,
  author = {S. F. Bhat and I. Alhashim and P. Wonka},
  title = {AdaBins: Depth Estimation Using Adaptive Bins},
  booktitle = {Proceedings of the IEEE/CVF Conference on Computer Vision and Pattern Recognition},
  year = {2021},
  pages = {4009--4018},
  url = {https://openaccess.thecvf.com/content/CVPR2021/html/Bhat_AdaBins_Depth_Estimation_Using_Adaptive_Bins_CVPR_2021_paper.html}
}

@inproceedings{Ref50,
  author = {A. Agarwal and C. Arora},
  title = {Attention Attention Everywhere: Monocular Depth Prediction with Skip Attention},
  booktitle = {2023 IEEE/CVF Winter Conference on Applications of Computer Vision (WACV)},
  address = {Waikoloa, HI, USA},
  publisher = {IEEE},
  month = {Jan},
  year = {2023},
  pages = {5850--5859},
  doi = {10.1109/WACV56688.2023.00581}
}

@article{Ref51,
  author = {J. M. Goo and X. Milidonis and A. Artusi and J. Boehm and C. Ciliberto},
  title = {Hybrid-Segmentor: Hybrid approach for automated fine-grained crack segmentation in civil infrastructure},
  journal = {Automation in Construction},
  volume = {170},
  pages = {105960},
  month = {Feb},
  year = {2025},
  doi = {10.1016/j.autcon.2024.105960}
}

@article{Ref54,
  author = {Z. Yu and C. Dai and X. Zeng and Y. Lv and H. Li},
  title = {A lightweight semantic segmentation method for concrete bridge surface diseases based on improved DeeplabV3+},
  journal = {Scientific Reports},
  volume = {15},
  number = {1},
  pages = {10348},
  month = {Mar},
  year = {2025},
  doi = {10.1038/s41598-025-95518-5}
}

@article{Ref55,
  author = {J. Zhang and Z. Zeng and P. K. Sharma and O. Alfarraj and A. Tolba and J. Wang},
  title = {A dual encoder crack segmentation network with Haar wavelet-based high–low frequency attention},
  journal = {Expert Systems with Applications},
  volume = {256},
  pages = {124950},
  month = {Dec},
  year = {2024},
  doi = {10.1016/j.eswa.2024.124950}
}

@inproceedings{Ref62,
  author = {L. Chen and X. Chu and X. Zhang and J. Sun},
  title = {Simple Baselines for Image Restoration},
  booktitle = {Computer Vision -- ECCV 2022},
  volume = {13667},
  series = {Lecture Notes in Computer Science},
  editor = {S. Avidan and G. Brostow and M. Ciss{\'e} and G. M. Farinella and T. Hassner},
  publisher = {Springer Nature Switzerland},
  address = {Cham},
  year = {2022},
  pages = {17--33},
  doi = {10.1007/978-3-031-20071-7_2}
}

@inproceedings{Ref63,
  author = {S. W. Zamir and A. Arora and S. Khan and M. Hayat and F. S. Khan and M.-H. Yang},
  title = {Restormer: Efficient Transformer for High-Resolution Image Restoration},
  booktitle = {2022 IEEE/CVF Conference on Computer Vision and Pattern Recognition (CVPR)},
  address = {New Orleans, LA, USA},
  publisher = {IEEE},
  month = {Jun},
  year = {2022},
  pages = {5718--5729},
  doi = {10.1109/CVPR52688.2022.00564}
}

@inproceedings{FADE,
  title={FADE: Fusing the Assets of Decoder and Encoder for Task-Agnostic Upsampling},
  author={Lu, Hao and Liu, W Emma and Fu, Hongtao and Cao, Zhiguo},
  booktitle={European Conference on Computer Vision (ECCV)},
  volume={13687},
  series={Lecture Notes in Computer Science},
  pages={231--247},
  year={2022},
  publisher={Springer Nature Switzerland},
  address={Cham},
  doi={10.1007/978-3-031-19812-0_14}
}

@inproceedings{CARAFE,
  title={CARAFE: Content-Aware ReAssembly of FEatures},
  author={Wang, Jiaqi and Chen, Kai and Xu, Rui and Liu, Ziwei and Loy, Chen Change and Lin, Dahua},
  booktitle={Proceedings of the IEEE/CVF International Conference on Computer Vision (ICCV)},
  pages={3007--3016},
  year={2019},
  doi={10.48550/arXiv.1905.02188},
  note={arXiv preprint arXiv:1905.02188}
}

@InProceedings{shit2021cldice,
    author    = {Shit, Suprosanna and Paetzold, Johannes C. and Sekuboyina, Anjany and Ezhov, Ivan and Johansson, Alexander and Lippert, Henning and Guibas, Leonidas J. and Menze, Bjoern H.},
    title     = {clDice - A Novel Topology-Preserving Loss Function for Tubular Structure Segmentation},
    booktitle = {Proceedings of the IEEE/CVF Conference on Computer Vision and Pattern Recognition (CVPR)},
    month     = {June},
    year      = {2021},
    pages     = {16560-16569}
}
\end{document}